\documentclass{article}
\usepackage{spconf,amsmath,graphicx, url}

\usepackage{enumitem}
\setlist{nosep, leftmargin=14pt}

\title{Effective Feature Learning for 3D Medical Registration via Domain-Specialized DINO Pretraining}

\name{Eytan Kats
\qquad Mattias P. Heinrich}
\address{Institute of Medical Informatics, University of Lübeck, Germany}

\begin{document}

\maketitle

\begin{abstract}
Medical image registration is a critical component of clinical imaging workflows, enabling accurate longitudinal assessment, multi-modal data fusion, and image-guided interventions. Intensity-based approaches often struggle with inter-scanner variability and complex anatomical deformations, whereas feature-based methods offer improved robustness by leveraging semantically informed representations. In this work, we investigate DINO-style self-supervised pretraining directly on 3D medical imaging data, aiming to learn dense volumetric features well suited for deformable registration. We assess the resulting representations on challenging inter-patient abdominal registration task across both MRI and CT modalities. Our domain-specialized pretraining outperforms DINOv2 model trained on large-scale collection of natural images, while requiring substantially lower computational resources at inference time. Moreover, it surpasses established registration models under out-of-domain evaluation, demonstrating the value of task-agnostic yet medical imaging–focused pretraining for robust and efficient 3D image registration. Our code and pre-trained models are publicly available at \url{https://github.com/EytanKats/med_ssl_3d}.
\end{abstract}


\section{Introduction}
\label{sec:introduction}

Image registration is a fundamental task in medical image analysis, playing a critical role in a wide range of clinical and research applications. By spatially aligning anatomical structures across different images or modalities, registration enables accurate longitudinal assessment of disease progression, supports multi-modal image fusion, and facilitates image-guided interventions. Reliable registration is therefore essential for quantitative analysis and informed clinical decision-making.

Despite its importance, medical image registration remains challenging due to several intrinsic factors. Variability in image intensities across scanners and modalities, as well as complex non-linear anatomical deformations, make direct intensity-based alignment unreliable. To address these challenges, modern approaches leverage robust feature representations that capture high-level semantic and structural information.

Hand-crafted descriptors, such as the Modality Independent Neighborhood Descriptor (MIND) \cite{heinrich2012mind}, have long been employed to extract discriminative and semantically meaningful features. Once feature maps are extracted, registration can be formulated as an optimization problem in feature space. Algorithms such as ConvexAdam \cite{siebert2024convexadam} estimate dense deformation fields that optimally align these features. These methods benefit from the semantic consistency of learned representations, enabling more robust registration compared to intensity-based optimization.

Pretraining models on large-scale datasets has recently demonstrated significant success in learning rich and transferable feature representations for a wide range of downstream tasks \cite{he2022masked, oquab2023dinov2, zhou2021ibot}. Models from the DINO family \cite{oquab2023dinov2} utilize self-distillation objectives to produce embeddings that capture both local and global semantic information. Recent studies \cite{song2025dino, gu2025vision} show that DINO features can be effectively applied to 3D volumetric registration without any fine-tuning. In these approaches, features are extracted from 2D slices and concatenated to form a 3D feature volume, which is subsequently projected to a lower-dimensional space using PCA \cite{halko2011finding}. The resulting features are then aligned using ConvexAdam \cite{siebert2024convexadam}. While these methods achieve promising registration accuracy, they require substantial upsampling of 2D slices to compensate for the relatively large patch size (14×14) of pre-trained DINO models, resulting in high computational costs and limiting their clinical applicability.

In this work, we investigate pretraining DINO directly on 3D medical imaging data from scratch, with the goal of improving zero-shot registration performance. Our main contributions are summarized as follows:
\begin{itemize}
\item We pretrain a transformer-based architecture using the self-distillation DINO framework on 3D medical data, incorporating design choices that enable computationally efficient inference for registration tasks.
\item We demonstrate that employing a smaller patch size is critical for achieving strong registration performance, as it preserves fine-grained structural details essential for accurate alignment.
\item We show that our pretrained model significantly outperforms DINOv2 and established registration models on challenging inter-patient abdominal registration tasks.
\end{itemize}

\begin{figure*}[t]
\centering
\includegraphics[width=\textwidth]{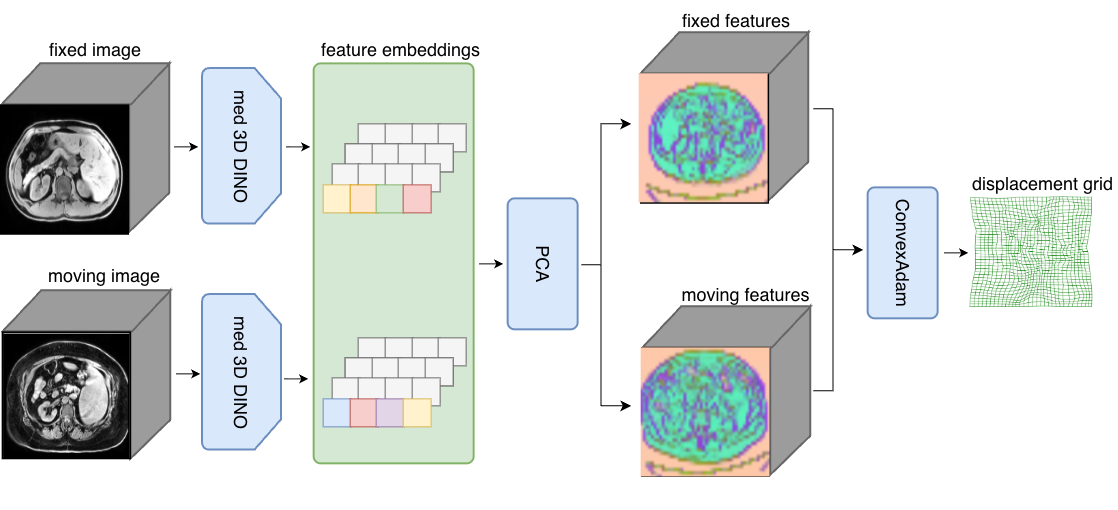}
\caption{Zero-shot registration framework. Features from the moving and fixed images are first extracted using the pretrained transformer and then projected into a low-dimensional space using PCA. ConvexAdam leverages these semantically rich representations to estimate an accurate displacement field.}
\label{fig:method}
\end{figure*}

\section{Method}
\label{sec:method}

Our approach consists of two stages. In the first stage, we pre-train a transformer-based model specifically designed for 3D medical images to learn dense, semantically rich feature representations in a self-supervised manner (Sec.\ref{subsec:pretraining}). In the second stage, the pre-trained feature extractor is employed in a zero-shot registration framework, where the learned representations guide the alignment of volumetric images without additional fine-tuning (Sec.\ref{subsec:registration}).

\subsection{Pretraining on 3D medical data}
\label{subsec:pretraining}
As the feature extractor, we employ the PRIMUS architecture \cite{wald2503primus}, an adaptation of the Vision Transformer (ViT) \cite{dosovitskiy2020image} specifically designed for 3D medical imaging tasks. For pretraining, the convolutional upscaling and decoding layers originally present in PRIMUS are removed, as our goal is to learn task-agnostic feature representations. The features are extracted from the output of the final transformer block, providing rich volumetric embeddings that capture spatial context.

Following the DINOv2 self-distillation framework \cite{oquab2023dinov2}, token embeddings are projected through a multi-layer perceptron (MLP) head to obtain high-dimensional feature vectors, upon which three complementary loss functions are jointly optimized to promote the emergence of semantically meaningful representations. The self-distillation loss aligns the projected class-token outputs between a student and a momentum-updated teacher network under different data augmentations, encouraging invariance to appearance and viewpoint changes. The iBOT loss \cite{zhou2021ibot} enforces local consistency between masked and unmasked patch tokens, enhancing spatial sensitivity. Finally, the KoLeo loss \cite{sablayrolles2018spreading} acts as a diversity regularizer, maximizing the spread of feature representations on the hypersphere to prevent collapse and ensure uniform coverage of the embedding space.

Together, these objectives guide the model toward learning volumetric representations specialized for the medical domain that capture both global semantic context and fine-grained anatomical details. These properties support accurate and robust dense correspondence estimation in downstream medical image registration task.

\subsection{Zero-shot registration}
\label{subsec:registration}

For registration, the features extracted from the final transformer block of the pretrained model are first projected into a low-dimensional space using PCA \cite{halko2011finding}, preserving the dominant components. These semantic feature volumes serve as input to the ConvexAdam algorithm \cite{siebert2024convexadam}, which estimates the dense displacement field between fixed and moving images. An overview of the zero-shot registration pipeline is illustrated in Fig.\ref{fig:method}.

ConvexAdam operates in two complementary stages, both of which depend critically on the quality of the feature representation. In the coupled convex stage, an initial deformation field is obtained by maximizing local feature correlations, yielding a robust coarse alignment. Building upon this initialization, the continuous refinement stage performs iterative optimization using Adam to minimize a sum-of-squared-differences (SSD) objective. This refinement progressively improves alignment while enforcing deformation smoothness through regularization. 

As both stages rely on the distinctiveness and semantic consistency of the feature embeddings, the overall effectiveness of ConvexAdam is tightly coupled to the quality of the learned representations. High-quality semantic features are therefore essential for achieving accurate and reliable registration.

\begin{figure}[t]
\centering
\includegraphics[width=0.46\textwidth]{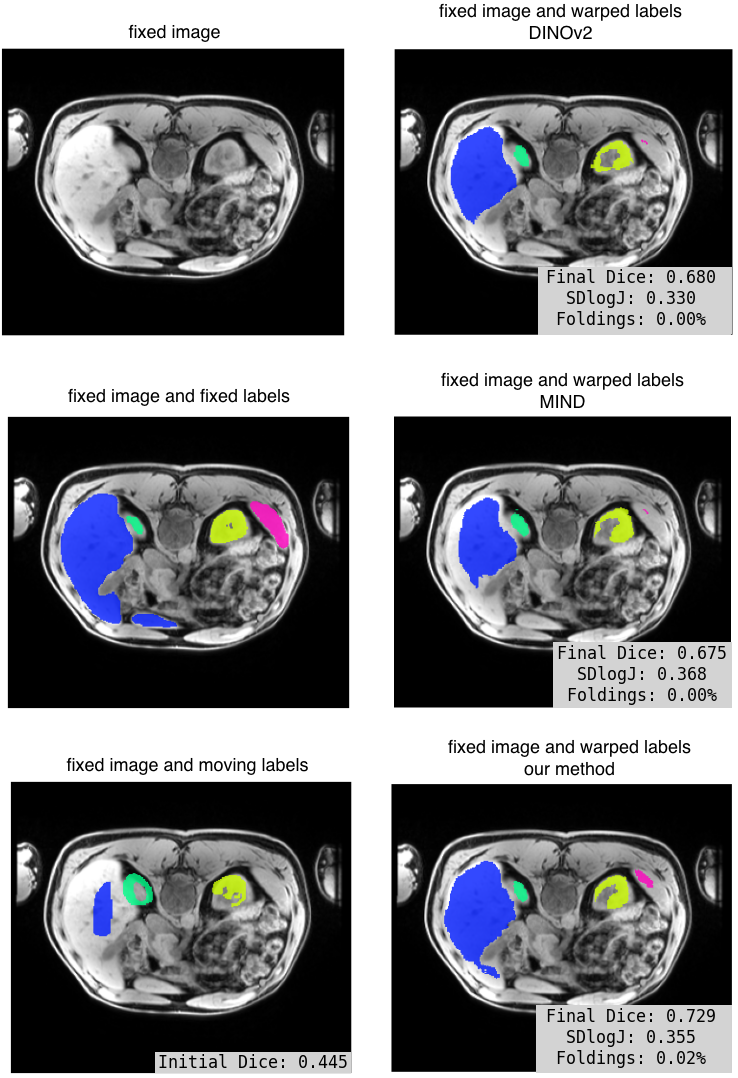}
\caption{Comparison of registration performance across different feature extractors. The results highlight the advantage of domain-specific DINO-style pretraining over DINOv2 and MIND features.}
\label{fig:results}
\end{figure}

\section{Experiments}
\label{sec:experiments}

We pretrain the feature extractor on a diverse collection of volumetric medical images, enabling the model to learn semantically rich feature representations. The learned representations are assessed in a zero-shot registration setting by applying the ConvexAdam algorithm, allowing to directly measure how well the pretrained features support volumetric images alignment without any additional task-specific fine-tuning.

\subsection{Datasets and evaluation metrics}
\label{subsec:data}

Our pretraining dataset comprises 1000 whole-body MRI scans from the German National Cohort (NAKO) study \cite{bamberg2022whole} and 1228 CT scans from the TotalSegmentator dataset \cite{wasserthal2023totalsegmentator}. Prior to training, all volumes undergo standardized intensity preprocessing. For MRI scans, intensities are first clipped to the 0.01\textsuperscript{th} and 99.9\textsuperscript{th} percentiles to suppress outliers, and then linearly scaled to the $[0,1]$ range. For CT scans only min-max scaling is applied.

We evaluate registration performance on challenging inter-patient abdominal datasets for both CT and MRI modalities. For CT, we use 10 validation scans (45 image pairs) from the AbdomenCTCT task of the Learn2Reg challenge \cite{hering2022learn2reg}, which include manual annotations of 13 anatomical structures. All volumes are resampled to an isotropic voxel spacing of 2\,mm and standardized to spatial dimensions of $192 \times 160 \times 256$ voxels. For MRI, we use a hold-out set of 10 abdominal scans (45 image pairs) from the NAKO dataset, annotated with 5 anatomical structures. These volumes are resampled to 1.5\,mm isotropic resolution and standardized to dimensions of $304 \times 256 \times 144$ voxels.

To quantify registration accuracy, we compute the average Dice similarity coefficient ($DSC$) over the available anatomical labels. Deformation regularity is assessed using two standard metrics: the standard deviation of the logarithm of the Jacobian determinant ($SD\log J$), and the percentage of voxels with negative Jacobian determinant ($|J|_{<0}$), which reflects the proportion of non-invertible (folding) deformations.

\subsection{Implementation Details}
\label{subsec:implementation}

Pretraining is performed for 150{,}000 steps using a batch size of 4. Most training hyperparameters follow the DINOv2 pretraining protocol \cite{oquab2023dinov2}. The base learning rate is set to $6 \times 10^{-5}$ with a linear warm-up over the first 500 steps. The student temperature is fixed at $0.1$, while the teacher temperature is linearly increased from $0.04$ to $0.07$ during the first 50{,}000 steps. Separate projection heads are used for the teacher and student networks, each mapping token embeddings to 8{,}192-dimensional feature vectors. The teacher network is updated using an exponential moving average (EMA) of the student weights, with the EMA momentum following a cosine schedule from $0.992$ to $1.0$ over the course of training.

Training is conducted on randomly sampled 3D patches of size $128 \times 128 \times 128$. We apply intensity augmentations as well as random 3D rotations. Masking of tokens follows a random strategy, with a masking ratio sampled uniformly between $0.4$ and $0.7$.

For zero-shot registration, ConvexAdam hyperparameters are tuned separately for each dataset and each feature extractor. Tuning is performed using an auxiliary set of 10 image pairs (not included in the main evaluation) by maximizing Dice-based alignment while preserving deformation-field smoothness.

\subsection{Results}
\label{subsec:results}

We first evaluate the impact of token embedding resolution on downstream registration performance (Tab.~\ref{tab:patch_size}). High-resolution tokenization proves essential for producing accurate deformation fields: the model pretrained with a patch size of $4 \times 4 \times 4$ achieves the strongest zero-shot registration results. This observation aligns with prior findings for hand-crafted MIND features \cite{heinrich2012mind}, which also rely on highly local neighborhoods to capture fine-grained structural information. The principal drawback of using such high-resolution tokens is the increased computational cost during pretraining, as the number of tokens grows cubically with spatial resolution. To mitigate inference-time cost, we configure the transformer with a lightweight architecture consisting of 8 encoder blocks and 6 attention heads, which reduces computation burden while maintaining strong representational quality.

\begin{table}[htb]
    \centering
    \resizebox{7.0cm}{!}{
    \begin{tabular}{c|c|c|c}
     patch size & $DSC$ & $SD\log J$ & $|J|_{<0}$ \\ \hline
     $12 \times 12 \times 12$ & ${49.09}_{\pm15.72}$ & $\textbf{0.264}_{\pm0.049}$  & $\textbf{0.01}_{\pm0.05}$\\
     $8 \times 8 \times 8$    & ${51.02}_{\pm17.21}$ & $0.269_{\pm0.059}$ & $\textbf{0.01}_{\pm0.04}$ \\
     $4 \times 4 \times 4$    & {$\textbf{57.74}_{\pm20.14}$} & $0.388_{\pm0.099}$ & $0.11_{\pm0.2}$ \\
    \end{tabular}}
    \caption{Impact of token resolution on downstream registration task. Increasing token resolution yields consistent performance gains, highlighting its role in capturing fine anatomical details.}
    \label{tab:patch_size}
\end{table}

Next, we compare the representation strength of our specialized medical DINO-style task-agnostic pretraining with two baselines on both the NAKO and AbdomenCTCT datasets: hand-crafted MIND features \cite{heinrich2012mind}, which are explicitly designed for registration, and DINOv2 \cite{oquab2023dinov2}, a self-supervised model trained on a large-scale dataset of natural images (Tab.~\ref{tab:methods_comparison}). For the widely adopted AbdomenCTCT registration task, we additionally include the results of TotalRegistrator \cite{pham2025totalregistrator} and UniGradIcon \cite{tian2024unigradicon} as reported in their respective publications, using versions that were not trained on AbdomenCTCT to ensure a fair comparison. Both TotalRegistrator and UniGradIcon were pre-trained on volumetric medical images specifically for registration, aiming to generalize effectively to unseen data.

\begin{table}[htb]
    \centering
    \resizebox{8.5cm}{!}{
    \begin{tabular}{c|ccc|ccc}
        & \multicolumn{3}{c|}{\textbf{AbdomenCTCT}} 
        & \multicolumn{3}{c}{\textbf{NAKO}} \\ 
        \cline{2-7}
        \textbf{method} 
        & $DSC$ & $SD\log J$ & $|J|_{<0}$ 
        & $DSC$ & $SD\log J$ & $|J|_{<0}$ \\ \hline

        Initial         
        & {$25.31_{\pm6.89}$} & - & - 
        & {$29.57_{\pm14.66}$} & - & - \\

        MIND \cite{heinrich2012mind}         
        & {$37.08_{\pm9.1}$} & $\textbf{0.451}_{\pm0.092}$ & $\textbf{0.29}_{\pm0.54}$ 
        & {$50.55_{\pm20.6}$} & $0.398_{\pm0.112}$ & $0.11_{\pm0.3}$  \\ 

        DINOv2 \cite{oquab2023dinov2}        
        & {$34.73_{\pm7.99}$} & $0.485_{\pm0.106}$ & $0.67_{\pm1.25}$ 
        & {$54.95_{\pm16.07}$} & $\textbf{0.326}_{\pm0.085}$ & $\textbf{0.01}_{\pm0.03}$\\

        TotalRegistrator \cite{pham2025totalregistrator}        
        & 36.2 & - & - 
        & - & - & - \\

        UniGradIcon \cite{tian2024unigradicon}        
        & 34.1 & - & 0.02 
        & - & - & - \\

        ours                                 
        & {$\textbf{39.76}_{\pm9.17}$} & $0.545_{\pm0.09}$ & $0.38_{\pm0.59}$
        & {$\textbf{57.74}_{\pm20.14}$} & $0.388_{\pm0.099}$ & $0.11_{\pm0.2}$
    \end{tabular}}
    \caption{Registration results on the AbdomenCTCT and NAKO datasets. The findings demonstrate the benefit of domain-specific DINO-style pretraining for dense deformable registration. Reported values for TotalRegistrator and UniGradIcon are taken from their respective publications and correspond to out-of-domain evaluations, where the models were not trained on AbdomenCTCT.}
    \label{tab:methods_comparison}
\end{table}

Because DINOv2 is a 2D model, features are extracted slice-by-slice and then stacked to form a 3D feature volume. Furthermore, DINOv2 was pretrained with a relatively large patch size of $14 \times 14$, which necessitates resizing each input slice to a high spatial resolution in order to produce features suitable for downstream registration. In our experiments, we used a feature map resolution of $64 \times 64$ and resized the slices accordingly. Larger resolutions were not evaluated due to excessive GPU memory requirements.

The results in Tab.~\ref{tab:methods_comparison} highlight the clear advantage of domain-specialized pretraining over the DINOv2 model trained on natural images. Compared to DINOv2, our model requires substantially fewer resources for inference (approximately 5GB of VRAM versus 35GB for DINOv2 on NVIDIA A100 GPU) and operates with a single forward pass, reducing computation time. Moreover, our approach outperforms established registration models, such as TotalRegistrator and UniGradIcon, which were explicitly trained for the registration task.

\section{Conclusions}
\label{ssec:conclusions}

Our study demonstrates that domain-specific DINO-style pretraining for 3D medical images yields superior feature representations compared to DINOv2 pretrained on natural images and outperforms established registration models. Importantly, our model requires modest computational resources during inference, enabling practical deployment in clinical settings.

The pretraining in our experiments is performed on a relatively large yet still limited collection of 3D medical volumes. Future work may involve extending the pretraining to larger and more diverse datasets, as well as adapting the approach to multi-modal registration scenarios, thereby further enhancing its generalizability and utility in clinical imaging applications.

\section{Compliance with ethical standards}
\label{sec:page}

The German National Cohort (NAKO) study is performed with the approval of the relevant ethics committees, and is in accordance with national law and with the Declaration of Helsinki of 1975 (in the current, revised version).

\section{Acknowledgments}
\label{sec:acknowledgements}

We gratefully acknowledge the financial support by German Research Foundation: DFG, HE 7364/10-1, project number 500498869.

\bibliographystyle{IEEEbib}
\bibliography{strings}

\end{document}